\documentclass[11pt,a4paper]{article}
\usepackage[hyperref]{acl2020}
\usepackage{times}
\usepackage{latexsym}
\usepackage{avm}
\usepackage{istgame}
\usepackage{dialogue}

\usepackage{microtype}
\usepackage{amsmath}
\usepackage{lingmacros}

\title{Different Games in Dialogue: Combining character and conversational types in strategic choice}

\author{Alafate Abulmiti \\
     Universit\'e de Paris, CNRS\\ 
       Laboratoire de Linguistique Formelle
    \texttt{alafate.abulimiti@gmail.com} }

\date{}
\aclfinalcopy 
\begin{document}
\maketitle

\begin{abstract}
    In this paper, we show that investigating the interaction of \emph{conversational type}
    (often known as language game or speech genre)
    with the character types of the interlocutors is worthwhile. We present a method of calculating the decision making process for selecting dialogue moves that combines character type and conversational type. We also present a mathematical model that illustrate these factors' interactions in a quantitative way.
\end{abstract}

\section{Introduction}
\citet{wittgenstein1953philosophical,bakhtin2010speech} introduced language games/speech genres as notions tying diversity of linguistic behaviors
to activity. Building on this, and insights of pragmaticists such as \citet{hymes1974ways,allwood-activity}, and earlier work in AI by \citep{allen-perrault80,cohen-perrault79}
\citet{larsson2002issue,ginzburg2012interactive,wongconversational} showed how to associate global structure with conversations in a way that captures the range of possible topics and idiosyncratic moves. \citep{larsson2002issue} is also the basis for an approach to building spoken dialogue systems (SDS) which is essentially domain general, offers a fine-grained treatment of grounding interaction, and which was extended to clarification interaction in \citep{purver2004theory}.

This body of work does not address, however the issue of strategic choice in conversation, which is the core issue underlying work in Game Theory. \citet{asher2017message} used game theoretic tools to develop a theory of strategic choice for dialogue. Although there are a variety of interesting insights captured in this approach, it is based on two assumptions that apply only to a restricted class of language games---games continue indefinitely and there exists a jury that assigns winning conditions to participants.

We seek to develop an approach to strategic choice applicable to the general case of dialogical interaction, where termination is an important consideration and where assessment is internal to the participants. Strategic choice is modelled by combining structure from conversational types with psychological and cognitive notions associated with the players.

Character type as a relatively independent factor abstracted out of the conversational type is important for dialogue. Although there is some analysis concerning both character effects and conversational types in the dialogue, combining them and analyzing their interactions in a quantitative way has not, to our knowledge, been carried out before. The purposes of this paper is, hence, to combine character type effect and conversational type analysis to yield a method that could help to analyse strategic choice in dialogue. 

\section{Background}

The starting point of \citep{asher2017message} is the framework of Banach-Mazurckiewicz games. They modify this framework to make it more amenable to analyzing certain kinds of NL dialogue, the emergent framework being \emph{BM messaging games}.
\citet{asher2017message} argued that each dialogue potentially continues indefinitely and has a winner adjudged by a third party jury. This is useful for modelling political discourse between rival groups or individual contenders in the public domain.
But clearly this sort of conception is not appropriate for a variety, arguably most types of dialogue.\footnote{Strictly speaking, \citet{asher2017message} allowed also for finite games, by allowing for games to consist of a special move from a certain point onwards, but this would seem to either defeat the purpose of assuming potential extendibility or to be purely formal.} These have beginnings ($InitState$) and a variety of distinct terminations ($FinState$) \cite{wong2018classifying}, and there is no `external jury' in most cases.


\citet{burnett2019signalling} developed a formal model called \textit{social meaning games} which explains how social variants affect the linguistic style of dialogue participants. And conversely, how the speaker's intrinsic linguistic style affects dialogue moves. \citet{pennebaker1999linguistic} shows that linguistic style is an independent and meaningful way of exploring personality.
There is evidence that people's personality traits influence their use of language. For instance, extroverted individuals are able to deceive others more easily, while neurotic individuals are less likely to be deceived \cite{riggio1988personality}. The charisma of a speaker has been shown to be closely related to an extroverted character \cite{bono2004personality}. There is also a strong relation between extroversion and conscientiousness and positive influences, as well as between neuroticism and dissent and various negative influences\cite{watson1992traits}. Thus, an individual's personality does affect decision-making process in dialogue.

Cooperation in dialogue is a widespread phenomenon, and \citet{allwood2000cooperation} identified four features of cooperation: cognitive consideration, joint purpose,
ethical consideration and trust. When engaging in a collaborative dialogue, where the interlocutor decides his next move based on the intentions of the other and a variety of information deriving from the context of the dialogue, it seems that character has a broad influence on the course of the dialogue. Thus, it seems natural that a
dialogue participant (DP) should also take into account the other's character traits in order to choose the appropriate move.

In the next section, we will explain the method we propose of combining character type effects and conversational
type.

\section{Methodology}
In this section, we wish to explore the interaction between character type and conversational type, by considering how given a possible range of moves, a quantitative analysis can be provided for move selection.

\subsection{Character Type}
 Researchers have developed a relatively unanimous consensus on personality description patterns, proposing the Big Five model of personality\cite{goldberg1992development}. 
Within this model there are  five traits that can be used to capture many aspects of human personality. The Big Five Personality (OCEAN) can be assessed by the NEO-PI-R\cite{costa2008revised}. The traits can be described as follows:

\begin{itemize}
    \item Openness: the ability to be imaginative, aesthetic, emotional, creative, intelligent, etc.
    \item Conscientiousness: displays characteristics such as competence, fairness, organization, due diligence, achievement, self-discipline, caution, restraint, etc.
    \item Extroversion: exhibits qualities such as enthusiasm, sociability, decisiveness, activity, adventure, optimism, etc.
    \item Agreeableness: the qualities of trust, altruism, straightforwardness, compliance, humility, empathy, etc.
    \item Neuroticism: difficulty balancing emotional traits such as anxiety, hostility, repression, self-awareness, impulsivity, vulnerability, etc.
\end{itemize}

\citet{goldberg1992development} gave a pertinent method to quantify character types in terms of 5 dimensional vector $[o, c, e, a, n]$. We define $\chi_s$ for the self character type scale vector and $\chi_o$ represents other character type scale vector.

In addition, with the development of machine learning and deep learning methods within NLP, a variety of approaches have been implemented for automatic recognition of
personality in conversation and text \cite{mairesse2007using}. \citet{jiang2019automatic} used attentive network and contextual embeddings to detect personality traits from monologues and multiparty dialogues. Given a text(or utterance), one can calculate the character type scale vector of this sentence with a robust prediction model. We define $c_i$ as $i$th dialogue move vector prediction.

We note that by calculating the similarity between the $\chi$ and $c_i$, we obtain the  extent to which a given dialogue move can fit either the self character type or the other character type. Note that $\chi_s$ is a dialogue interlocutor's intrinsic property which does not show great change during conversation, but considering one's imperfect information situation, $\chi_o$ will change once new evidence arises and can be modified  by applying Bayes' rule. 

\subsection{Conversational Type}
\citet{pennebaker1999linguistic} also indicated that linguistic style gets influenced by the situation in which the interlocutor find themselves. \citet{wongconversational} provided a topological perspective on the space of conversational types based on the distribution of Non-Sentential Utterances (NSU) within each type. \citet{wong2018classifying} developed a model of a conversational type couched in TTR
\cite{cooper2005records}. On this view, a conversational type is a 4-tuple  \{$ConvRules$, $InitState$, $FinState$, $G$\}, here $ConvRules$ represents a set of conversational rules, transition rules between different dialogue states (\emph{dialogue gameboards}) of type $DGBType$ \citep{ginzburg2012interactive}, $InitState$ and $FinState$ are the initial and final DGB states. \\
$DGBType$ $\mapsto$\\
\begin{avm}
    \[
        spkr: Ind \\
        addr: Ind \\
        utt-time: Time \\
        c-utt: adressing(spkr, addr, utt-time) \\
        FACTS: set(Proposition) \\
        Pending: list(Locutionary Proposition) \\
        Moves: list(Locutionary Proposition) \\
        QUD: poset(Question) \\
    \]
\end{avm}

G is the grammar which serves for the different conversational language use.

\paragraph{An Example}
Considering a commercial transaction conversational type involving shopping in a bakery, so that we have: \\
$Bakery =$ \\
\begin{avm}
    \[
        participant: InteractionGroup \fbox{\d{$\wedge$}}
        \[
            c1:customer(A) \\
            c2:baker(B) \\
        \] \\
        qnud-set = QS : poset(question) \\
        c1 :
        \{$\lambda$ x.InShopBuy(A,x), $\lambda$ P.P(A),\\
        $\lambda$ P.P(B), $\lambda$ x.Pay(A,x)
        \}
        $\in$ QS\\
        moves : list(IllocProp)
    \]
\end{avm}

this involves:
\begin{itemize}
    \item participants: Baker and customer.
    \item qnud-set: questions to be discussed during the interaction
    \item moves: Dialogue moves made.
\end{itemize}

\citet{wong2018classifying} argued that clarification interactions provide data showing that interlocutors can be uncertain about the conversational type that classifies the interaction they are in, as for instance in (\ref{ctcr}):

\eenumsentence{\item[] \label{ctcr} (Context: A is being interviewed by B)  A: Hi B: Hi . . . B (1): have you seen Blackklansman yet? A (2): Wait— is this an informal chat or a formal interview? B: A bit of both. (Example (54) from \citep{wong2018classifying}.
}

Thus, the DP has an opaque or uncertain "guess" as to the conversational type and this also influences the decision-making process. Following probabilistic TTR theory \cite{cooper2014probabilistic}, we assume that the DP has a probabilistic conversational type in the private part of information state and this can be updated during dialogue by Bayesian inference. We model the information state as follows:\\
$InformationState =$\\
\begin{avm}
    \[
        private = \[
            CharacterType:\[
                Self: Vector \\
                Other: Vector \\
            \] \\
            Goals: Set(Prop) \\
            Tmp: private \\
            ConvType : ConvType \\
            Conv-prob : [0, 1]
        \] \\
        dgb :DGBType
    \]
\end{avm}

In the private part, we have the \textit{CharacterType} stores for self and other's character type vector; \textit{Goals} tracks a set of (futurate) propositions that the DP wants realized;
\textit{Tmp} is a backup of the previous private state. Its use  is for reasoning  about the current \textit{CharacterType} and updating the \textit{Conv-prob}; \textit{ConvType} is the conversational type and here we introduce \textit{Conv-prob}, which indicates the probabilistic conjecture concerning the current conversational type.

\subsection{Move Space}
Individuals choose the moves from different kind of possible options. \citet{ginzburg2019characterizing} provided a taxonomy of the response space of queries. 14 possible categories are given by the study of English and Polish corpus. The field of Nature Language Generation has some useful probabilistic methods to generate the range of potential moves, such as the Sequence-to-Sequence model\cite{sutskever2014sequence}. \citet{see2019makes} proposed a new conditional weight control technique to make the response space more human-like. The popular pre-training model like BERT\cite{devlin2018bert} exemplifies NLG capability with fine-tuning. We define here $A$ for a dialogue move space vector composed by $a_i$ representing the $i$th move.

In the current work we will not discuss how to specify the possible moves, but will leave this for future work.

\subsection{Decision Making}
\subsubsection{Global View}
\citet{levelt1993speaking} proposed that speakers monitors themselves while speaking. In other words, individuals have a self-criticism mechanism enabling them to reflect on their behaviors, emotions and speaking. We dub this the \textit{SelfMonitor}.

Given our analysis in the first two subsections, a DP's move choice in the dialogue is influenced by her character type, the other interlocutor's character type, and the conversational type. When a DP tries to respond to the other party's dialogue move, she first constructs a dialogue move space, which yields a set of possible utterances that the DP can use. The DP typically makes a conjecture about the other's character type in terms of individual personality traits, based on her a priori knowledge of that individual  and the current state of the dialogue. In addition, the DP has a probabilistic assumption about the present conversational type given her cognitive state. Subsequently, the DP's \textit{SelfMonitor} determines which factor is more valuable in this context. Eventually a move is selected considering each possible move's value by comparing the affinity of each move for each factor with the skewness of each factor as determined by the \textit{SelfMonitor}.

\subsubsection{Mathematical Modeling}
After the above analysis, we offer a mathematical model to explicate this process. In evaluating possible moves, we have three important factors: self character type, other character type, and the conversational type. We want to provide a real valued function $\rho$  to evaluate each move in the dialogue move space.

\begin{itemize}
    \item $a_i$: $i$th move.
    \item $\overrightarrow{\chi_s}$: Self character type vector.
    \item $\overrightarrow{c_i}$: character type vector for $i$th move.
    \item $\alpha$ Weight for the self character type effect.
    \item $\overrightarrow{\chi_o}$: Current other character type vector estimation.
    \item $\beta$: Weight for the other character effect.
    \item $p$: Probabilistic conjecture of the conversational type.
    \item $d_i$: $i$th move conformity with the conversational type from -1 to 1.
    \item $\gamma$ Weight for probabilistic conversational type.
    \item $W = [\alpha, \beta, \gamma]$.
\end{itemize}

Conformity represents the degree to which a dialogue move conforms to the current conversational type. In other words, it can be modeled as the evaluation score generated by this dialogue move in dialogue context.

In order to calculate the "affinity" among character type vector \{$\chi_s$,$\chi_o$\} and move vector ($c_i$), we use \textit{cosine} similarity, defined as follows:
\begin{equation}\label{1}
    simi(A, B) =\cos (\theta)=\frac{A \cdot B}{\|A\|\|B\|}
\end{equation}
Then we define the function $\rho(a_i)$:
\begin{equation}\label{2}
    \rho(a_i)=\alpha \cdot simi(\overrightarrow{c_i}, \overrightarrow{\chi_s}) + \beta  \cdot simi(\overrightarrow{c^i}, \overrightarrow{\chi_o}) + \gamma  d_i \cdot p \\
\end{equation}
and let $X=[simi(\overrightarrow{c_i},\overrightarrow{\chi_s}),\ simi(\overrightarrow{c^i}, \overrightarrow{\chi_o}),\ d_i \cdot p]^T $
as a decision factor matrix, obtaining:
\begin{equation}\label{3}
    \rho(A)=W \cdot X
\end{equation}

where $\alpha+\beta +\gamma =1$


$\alpha$, $\beta$, $\gamma$ are in fact estimated by the \textit{SelfMonitor} based on information deriving from the information state. We believe those weights are mainly fixed in the beginning of the conversation, because great changes in strategic choice lead to the suspicion of deception in some cases \cite{riggio1988personality}, This estimation process can be back-engineered by observing DP's selection of moves in the dialogue.


After the calculation of $\rho$, we get the score for each move. This score alone cannot determine the final decision---we need to take into account the features that we have not yet discussed and observed, so here we will probabilize them, i.e., convert it into a probabilistic distribution space using the $softmax$ function.

\begin{equation}\label{4}
    softmax(a_i)= \frac{exp(\rho(a_i))}{\sum_{j}^{ }exp(\rho(a_j)))}
\end{equation}

We then obtain a probability estimate for each move, which indicates that the greater the probability the DP is more inclined to choose this move.

\subsection{Example}
Here we illustrate our proposed approach with an example.
\subsubsection{Scenario}
In a bakery, we observe a customer and baker's buying and selling process during the pandemic of COVID-19.
\paragraph{Goals}
For simplicity we fix the final goals as follows:

\eenumsentence{
    \item Customer goal: buy two croissants.
    \item Baker goal: sell the two croissants and obtain the desired price.
}

It is worth noting that both players may have more complex goals or a series of goals. For example, the customer might want to have appropriate desserts for dinner. In order to achieve those kind of goals, we often divide into several specific and simple goals. In such a case, the sub-goal might be:

\begin{enumerate}
    \item what are the best desserts in this bakery?
    \item From among those best desserts, which one will fit the dishes I prepare tonight?
\end{enumerate}

For our current purposes, however, we will avoid such complications, despite their importance in a variety of cases. 

\subsubsection{Move Space}
We  assess a baker's possible responses to the customer's initial utterance: "2 croissants". We assume the following possible responses of the baker:
\eenumsentence{\item[]
    \begin{dialogue}
        \speak{client} 2 croissants.
        \speak{baker}
        (i) 1.90. \\
        (ii) Get out of the bakery, you're not wearing a mask. \\
        (iii) Please would be nice. \\
        (iv) 1.90 and please would be nice.
    \end{dialogue}
}
(i) would lead to a quick end that  meet their needs. This "style" is used in most cases in our life disregarding other conversational factors and attending only to the baker's final goals. If the baker considers particularly the conversational type's impact, he would choose (i) to advance the conversation, thereby looking forward to finalizing the conversation early and achieving his goal. 

(ii) would lead to an "unpleasant" conversation. The baker thinks that the customer's behavior is disrespectful (a short demand lacking politeness). He therefore uses the lack of the mask as a pretext or has in mind a competing goal for fighting against this disrespect. Finally, neither participants goals is achieved. 

(iii) This choice shows that the baker wants to have a pleasant and respectful conversation above all. It is clear that baker does not assign much weight to his final goals or the conversational type. Instead he prioritizes his psychological needs. The question under discussion would  shift to another topic and the conversation might evolve into a dispute, though with lower probability than in (ii). 

(iv) indicate that the baker wants to persist with the trade without disturbing his final goals and his psychological needs. This seems to be very much of a compromise move, but still the final state would depend on the customer's interpretation of the dialogue pragmatics.

\subsubsection{A worked example}
For this example, we have a character type vector of [o, c, e, a, n] we assume that baker character vector is $\chi_s$ = [0.0, 0.3, 0.0, 0.0, 0.5] showing conscientiousness and relatively high neuroticism and \textit{Conv-prob} is $p$ = 0.98 assuming the baker's conformity with the bakery conversational type is high. Following the initial state $S_0$, the baker hears "2 croissants" from client, after which the baker updates the information state then the \textit{SelfMonitor} chooses values for $\alpha, \beta, \gamma$. For example, we assume $\alpha$ = 0.1, $\beta$ = 0.1 and $\gamma$ = 0.8, we assume, in this circumstance, that the baker thinks more about the conversational type rule than self character type and other character type, then:

$$CharacterType.Other= \chi_o$$
$$\chi_o=\mu(\Uparrow Tmp.CharacterType.Other, \Uparrow dgb)$$

$\mu$ is the other's character type updating function with parameters: the previous state's character type vector and the current DGB.
We assume that after the other's character type update we have $\chi_o$ = [0.0, 0.0, -0.1, -0.4, 0.2] assuming the customer is not very agreeable (-0.4) and a little neurotic (0.2).

(i): We assume that "1.90" shows disagreeableness (-0.4) and slight neuroticism (0.2) so $c_1$ = [0.0, 0.0, -0.1, -0.4, 0.2] and $d_1$ = 0.8 shows "1.90" shows high conformity with the bakery conversational type. Then according to function (\ref{2})
we have: $\rho(a_1)$ = 0.7646

(ii): We assume that "Get out of the bakery, you're not wearing a mask" shows high disagreeableness (-0.7), low conscientiousness (-0.5) and high neuroticism (0.8) so $c_2$ = [0.3, -0.5, 0.0, -0.7, 0.8] and $d_2$ = -1.0 shows incongruity with the bakery type. As a result, we obtain $\rho(a_2)$ = -0.7080

(iii): We assume that "Please would be nice" shows high agreeableness (0.7) and slight extroversion (0.3) $c_3$ = [0.2, 0.0, 0.3, 0.7, -0.2] and $d_3$ = 0.3 shows low conformity with the conversational type. Consequently, we obtain $\rho(a_3)$ = 0.1201

(iv): We assume that "1.90 and please would be nice" shows relatively high openness (0.5), conscientiousness (0.6) and extroversion (0.4), high agreeableness (0.7) and low neuroticism (-0.4) so $c_4$ = [0.5, 0.6, 0.4, 0.7, -0.4] and $d_4$ = 0.7 shows high conformity with the bakery
type. So we obtain, $\rho(a_4)$ = 0.4727

We then apply those preference score with the $softmax$ function:
$softmax(\overrightarrow{\rho(a_i)})$ = [0.3998, 0.0917, 0.2099, 0.2986]

This indicates that the probability that the baker selects (i) is 39.98\% (iv) is the next ranked score. All this assuming that the baker's character type is relative high neuroticism and the commitment degree to the conversational type is high. Given his character type, it is reasonable to conclude that he would not be extremely polite (option iv) but not rude because of the constraint of conformity with  the conversational type. 

Now we want to illustrate that if we change the distribution of certain of the parameters, things can flip significantly.
We change the distribution of different variables ($\alpha,\beta,\gamma$) to [0.3, 0.1, 0.6] showing that the baker is concerned a little more about his own character type  (0.3 rather than 0.1) but slightly less than before  on conversational type (0.6).
And we assume the baker's character type vector is [0.5, 0.7, 0.3, 0.8, -0.5] which shows high conscientiousness (0.7), low neuroticism (-0.5) and high agreeableness. Then we obtain:$\rho(\overrightarrow{A})$ = [0.3457, -0.7277, 0.3217, 0.6946]. the resulting following with probabilistic distribution is $softmax(\overrightarrow{\rho(a_i)})$ = [0.2677, 0.0915, 0.2613, 0.3795]

This indicates that the baker would choose option (iv) for the next move with 38\% probability, an option which balances the baker's final goal and personal psychological needs (high conscientiousness).

Finally, we modify ($\alpha,\beta,\gamma$) to [0.8, 0.1, 0.1] and assume that the baker's character type vector is [0.2, -0.3, 0, -0.5, 0.8] which shows high neuroticism (0.8), low agreeableness (-0.5). What we obtain is
$\rho(\overrightarrow{A})$ = [0.8007, 0.7652, -0.5229, -0.5032]. From this what results is the following:
$softmax(\overrightarrow{\rho(a_i)})$ = [0.3996, 0.3856, 0.1064, 0.1085]

This indicates that the probability of choosing (i) and (ii) is now about the same. In this scenario the baker  focuses more on his own character type. (ii) shows complete incongruity with the bakery type, which indicates a complete violation of the baker's original final goal. Hence,  at this point, the baker is in a dilemma. It is worth noting that this state will lead to a "breakthrough point", and if the baker chooses (ii), it means that baker chooses to change his final goal or conversational type. The consequence of this is that the $Goals$, $ConvType$ and $Conv-prob$ should be changed in the private state of information state. How to effect this in a formal way, we leave to future work.

\section{Conclusion}
Character is a person's stable attitude towards reality and  it also affects one's performance in dialogue. Conversational type is, in one way or another, since the early days of AI work on dialogue, one of the principal notions of dialogue research. It reflects domain specific interaction, in particular the move space given to the conversational participants. We have tried to show in this paper that investigating the interaction of these two factors is worthwhile. In particular, we present a method of the decision making process for moves that combines character type and conversational type. We present a mathematical model that combines these factors after assigning numerical values to the various parameters that are involved and demonstrate by means of an example how this works.

\section{Future Work}
In this paper, we have made a preliminary proposal concerning the modelling of character types and their combination with conversational types. We aim to refine this in future work in ways that include the following:

\begin{itemize}
    \item \citet{wong2018classifying} gave a formal method to classify conversations into types. We believe that under realistic conditions, there are often multiple conversational types involved in a single conversation, which may involve sequential transformations or overlapping phenomena.
    \item In the approach sketched here, for classifying the conversational type we use a probabilistic TTR. However, in practice this assessment can change as the dialogue unfolds. We hope to develops methods that incorporate such dynamism.
    \item Personality analysis based on the Big Five theory is robust, but inference about each other's character types is also in flux during conversations, and examining the impact of the change process on our approach is worth looking forward to future work.
    \item \citet{ginzburg2019characterizing} provided a categorical approach to the response to the question. We hypothesize that the conversational type has the role of delimiting the range of possible moves. We aim to characterize this variability.
    \item This article introduces the concept of move conformity, which is required for automatic detection of research based on different types of dialogue in the future work. This could be achieved by modifying an existing NLG evaluation model (e.g. BLEU \cite{papineni-etal-2002-bleu}, ADEM \cite{lowe2017towards}).
    
    \item This paper discussed several cases in detail, but they are all constructed examples, so fitting this model on actual conversations (e.g.~the British National Corpus) or scripted dialogues annotated for character types   (e.g.~FriendQA \cite{yang2019friendsqa}) with experimental predictions is  desirable.
\end{itemize}

\subsection*{Acknowledgments} 

{\small This work was supported by an internship funded by the Institut Universitaire de France, within Jonathan Ginzburg's project {\it Unifying Verbal and Non-verbal Interaction}. We also acknowledge the support of the French Investissements d'Avenir-Labex
EFL program (ANR-10-LABX-0083). Thanks to Prof. Ginzburg's patient guidance during the internship. Thanks also to three anonymous reviewers for SemDial for their detailed and perceptive comments.
}
\bibliography{game}
\bibliographystyle{acl_natbib}
\end{document}